\documentclass{article}
\usepackage{spconf,amsmath,graphicx,url}

\title{Addressing speaker gender bias  in large scale speech translation systems}
\name{Shubham Bansal, Vikas Joshi, Harveen Chadha, Rupeshkumar Mehta, Jinyu Li}
\address{Microsoft Speech and Language Group}
\begin{document}
%
\maketitle
\begin{abstract}
This study addresses the issue of speaker gender bias in Speech Translation (ST) systems, which can result in offensive and incorrect translations. The masculine bias often found in large-scale ST systems is typically perpetuated through training data derived from Machine Translation (MT) systems. Our approach involves two key steps. First, we employ Large Language Models (LLMs) to rectify translations based on the speaker’s gender, in a cost-effective manner. Second, we fine-tune the ST model with the corrected data, enabling the model to generate gender-specific translations directly from audio cues, without the need for explicit gender input. Additionally, we propose a three-mode fine-tuned model for scenarios where the speaker’s gender is either predefined or should not be inferred from speech cues. We show  absolute 70\% improvement in translations for female speakers compared to our baseline and other large-scale ST systems, such as Seamless M4T and Canary on MuST-SHE testset.
\end{abstract}
\begin{keywords}
speech translation, gender bias
\end{keywords}

\section{Introduction and Related work}

Speech translation (ST) transforms spoken words to text in another language. They have applications in automatic video subtitling, dubbing, and facilitating cross-lingual communication. Traditionally, ST followed a cascaded approach. First, a speech recognizer \cite{zhang2020transformer, li2019improving} converted spoken words into text in the source language. Then, a machine translation (MT) model \cite{bahdanau2014neural, stahlberg2020neural} translated this text into the target language, referred to as \textit{cascaded ST}. Recent advancements introduced \textit{direct ST models}~\cite{berard2018end,gaido2020knowledge,xue2022large, xu2023recent, gandhareliterature}, which directly map speech to translated text. The direct ST models can perform real time speech translation, while the cascaded ST models need to wait for text in the source language to be available before translating it to the target language. The direct ST models are also easier to maintain, as a single model is trained and deployed. The direct ST models have slightly inferior translation quality but recent studies~\cite{bentivogli2021cascade} have shown that the gap  is reduced and direct ST models outperform cascaded ones on certain automated metrics~\cite{etchegoyhen2022cascade}. Despite their effectiveness, ST models are not immune to biases. \\
In this work, we specifically address gender bias exhibited by direct ST models. Gender bias \cite{koolen2017these, sun2019mitigating} is the unequal treatment or representation of genders and has also been studied extensively for several NLP and Speech tasks \cite{nadeem2020stereoset, savoldi2021gender,  bentivogli2020gender, gaido2020breeding, tatman2017gender}. It can influence how people interact and perceive others, resulting in discrimination and obstacles for minorities. Gender bias can arise in different linguistic levels, such as lexical, morphological and syntactic. For example, lexical bias occurs when gendered words are used inaccurately.
In the context of speech translation,  such gender bias manifests when the model generates translations dominantly in  masculine form. This may result in inaccurate translations, as the system fails to respect the speaker’s gender identity. It can also lead to offensive translations~\cite{bourguignon2015does}, perpetuating stereotypes and excluding those who identify as female from benefiting fully from these technologies. \\
Globally, a significant number of languages \cite{Aristar1992GrevilleCG} have grammatical gender, where nouns, pronouns and related words have gender distinctions, such as masculine or feminine forms. English, for instance, employs gender-specific pronouns like “he” or “she,” while other languages exhibit even more pronounced gender distinctions.
Consider the sentence “I am a teacher”. In Spanish, this simple phrase translated into two distinct forms: “Soy profesor” (for self-identified males) and “Soy profesora” (for self-identified females). The Spanish word for “teacher” adapts to the speaker’s gender. Similarly, when translating “I go to school” into Hindi, the phrase varies based on speaker gender. Therefore, when translating English speech into gender-specific languages, it’s crucial to account for the speaker’s gender and select appropriate words that align with the language’s gender-specific forms.

 Most machine translation (MT) models default to the masculine form as speaker gender cues are often absent in the English input text.
 \cite{saunders-byrne-2020-reducing, saunders-etal-2020-neural}  advocated for the refinement of MT models using small, intentionally crafted, gender-balanced datasets when gender cues are discernible in co-referential links, thereby reducing gender bias. Alternatively,~\cite{savoldi-etal-2021-gender, bentivogli2020gender} proposed to add a gender tag to the  input during training and inference, however, it is challenging to identify the text with gender specific forms for large-scale training data.~\cite{hassanawadalla2018gender} showed promising results with language-specific hand crafted rules to identify such text in Arabic, however, the rules does not generalize to other languages.
Most recently, the MuST-SHE WMT23 test results \cite{savoldi2023test} showed that the feminine accuracy were less than 30\%  for majority of the MT systems. Consequently, even when gender information is available, direct ST  models may inadvertently exhibit gender bias as they are trained from translations generated by MT models or automatic alignment techniques \cite{barrault2023seamless}, which may have gender bias. 

This work centers on tackling gender bias in English-to-Non English direct ST models 
 %
by exploiting speech cues~\cite{bentivogli2020gender}. However, it faces a challenge: the need for diverse training data containing accurate gender forms, which remains scarce.  There are studies which utilized MuST-C corpus to address gender bias in direct ST models\cite{bentivogli2020gender,gaido2020breeding}, however, the authors reported that masculine bias still existed. Hence, they proposed gender controlled translation that requires prior knowledge of speaker gender but gender accuracy was still reported to be less than 70\% for female speakers with a single model. The previous works\cite{bentivogli2020gender, gaido2020breeding} relied on human translated data such as MuST-C \cite{cattoni2021must} or augmented Librispeech \cite{kocabiyikoglu2018augmenting}. To the best of our knowledge, gender bias in large scale direct ST models is not studied yet. Our focus lies in reducing gender bias in large-scale direct ST models trained on tens of thousands of hours of data. The key contributions of this work are as discussed below. 

\textbf{Gender de-biasing for large scale training data:}   Obtaining human-annotated gender-debiased translations for tens of thousands of hours of training data is expensive. We use large language models (LLMs), more specifically GPT-4\cite{touvron2023llama, achiam2023gpt} to generate gender-debiased translations  in the ST training data. We adopt techniques such as few-shot prompting \cite{ma2023fairness}, chain of thought \cite{wei2022chain} and batch inferencing, to improve accuracy and reduce cost. Moreover, we do not rely on LLMs for the entire training data, but only  a subset of it, and show that weighted fine-tuning with such data is sufficient to adapt a large-scale ST model to generate gender specific translation while not degrading the BLEU \cite{papineni2002bleu} score.

\textbf{Respecting speaker's gender identity:}  Gaido et al.~\cite{gaido2020breeding} emphasized that speech cues are not a generalized solution for all users, such as transgender individuals and children.  We agree that perceptual markers are not comprehensive for gender identification and also recognize the importance of situations where gender is indeterminate or preassigned. Our work proposes to adapt the ST model architecture that can generate accurate speaker gender forms from audio inputs in an ‘Auto’ mode or allow the user to choose the desired speaker gender form in a ‘Masculine’ or ‘Feminine’ mode, respecting the diversity of speakers. 

\textbf{Gender representation (GR) loss:} We also propose to have an additional GR loss during ST model training 
and show that having GR loss improves on generating gender specific translations when inferred directly from speech.

We experiment with English to Spanish (ES) and Italian (IT)  ST model and show that our proposed approach achieves more than 90\% average gender translation accuracy on MuST-SHE testset.  We also compare state of the art ST models like Seamless M4T \cite{barrault2023seamless} and Canary \cite{blog} and show superior performance of our models.
\section{Types of gender bias}
Before delving into our methodology, it is crucial to outline the major categories of gender bias (as defined in latest MuST-SHE release), and specify the type of bias we aim to address:

\textbf{Category 1}: In instances where the speaker’s gender cannot be deduced from the English text (e.g., “I am a teacher”), but can be inferred from audio cues or explicitly shared by the speaker.  Our primary focus in this paper is on addressing bias within this category.


\textbf{Category 2}: When the gender of the speaker or another person is evident from the English text context (e.g., “She is a doctor”), but is inaccurately translated in language X. Although improving this type of bias is not our primary focus, we want to ensure that our approach does not regress on Category-2 bias. 

\textbf{Category 3}: In cases where a third person is referenced, and the gender of this third person is unknown from both the text and audio cues (e.g., “There is a doctor”), our objective is to preserve the original translation without altering the gender form of the translation.
\section{Method} 

Training data for direct speech translation models is often generated at scale by translating human-labeled speech recognition data using a MT model \cite{jia2019leveraging, xue2022large}. This approach is cost-effective compared to obtaining human-labeled translations of speech data. As most MT models exhibit gender bias, the training data for direct ST models also inherits this bias. Hence, we propose to first de-bias a subset of training data efficiently and subsequently fine-tune the ST model to automatically generate gender-specific translations, either based on the audio cues or the gender information provided by the speaker. 
\subsection{Gender de-biasing for large scale training data}
\begin{figure}[t]
\centering
\includegraphics[scale=0.5]{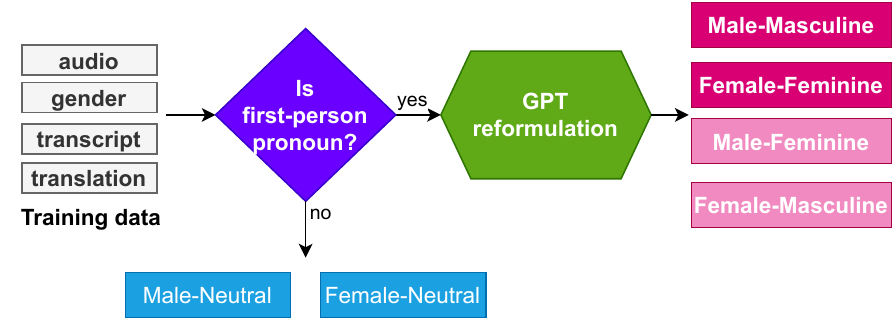}
\caption{Gender de-biasing for large scale training data.}
\label{fig:data_prep_fig} 
\end{figure}

Fig.~\ref{fig:data_prep_fig}  illustrates our approach to de-bias large-scale training data. Firstly, the input data undergoes filtering using data selection methods. Subsequently, we reformulate a subset of the data with GPT-4 using appropriate prompt. Finally, we prepare the targets in a suitable form for fine-tuning the ST model. Each of these steps is described in detail in the following sub-sections.

\subsubsection{Data selection} Not all English sentences yield gender-specific translations. For instance, phrases like “Play a song” or “How is the weather” produce gender-neutral translations. We hypothesize that English first-person pronouns, such as “I” and “my”, which are self-referential, are more likely to result in gender-specific translations. Consequently, we refine the training data to include only tuples (Audio, English Transcription, X Translation) that contain first-person pronouns in the English transcription, reducing our subset to 19.4\% of the original data. Our hypothesis is bolstered by the observation that 97.8\% of utterances in the MuST-She dev set, exhibiting Category1 gender bias, contain at least one first-person pronoun in the English transcription. Also, majority of our training data already includes speaker gender labels. If not, we ignore such utterances from our fine-tuning data. Further, we only sample 2 million utterances from this subset for GPT-4 reformulation, ensuring equal gender proportions. We will show that these 2 million utterances are sufficient to adapt the ST model to generate gender-specific translations, thereby significantly reducing the cost of GPT reformulation.

\subsubsection{Prompting methodology for GPT-4 reformulation}

\begin{figure}[htbp]
  \centering
  \includegraphics[width=\linewidth]{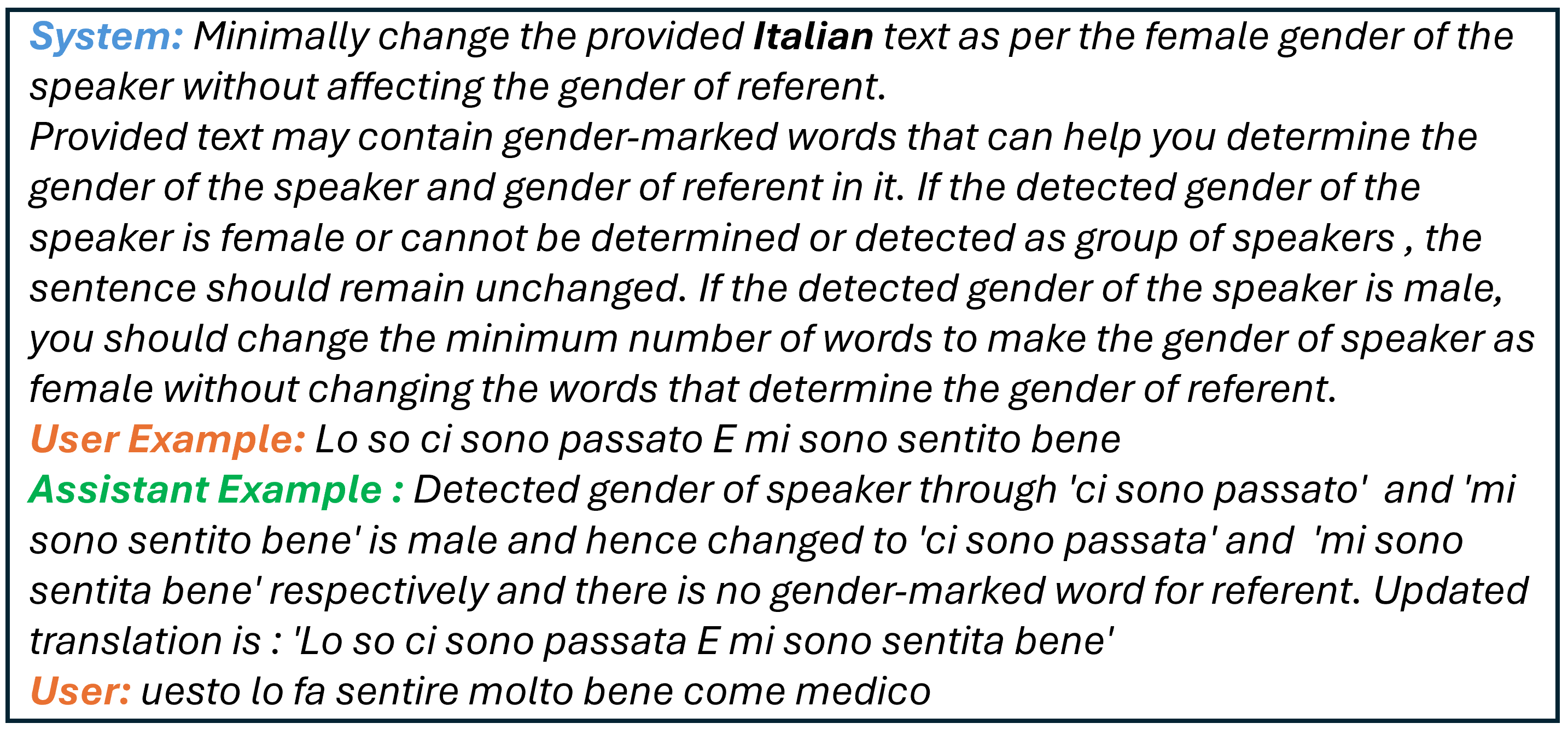}
  \caption{Prompt to obtain feminine translation form.}
  \label{fig:GPTprompt} 
\end{figure}

Gender systems significantly differ across languages, affecting the processes of gender assignment and agreement for gendered lexical items. Manipulating the gender alternation of the speaker, without affecting the referent, poses a challenge due to the need for language-specific, manually crafted rules. We suggest a universal method that employs few-shot~\cite{ma2023fairness} and chain-of-thought~\cite{wei2022chain} prompting techniques to harness the generative capabilities of GPT-4. Recent works \cite{jiao2023chatgpt,wang2023document} have also demonstrated that large language models (LLMs) such as GPT-4 can match the performance of commercial MT systems. Our approach involves instructing GPT-4 to generate both masculine and feminine versions for the speaker in each target language for 2 million utterances. Fig.~\ref{fig:GPTprompt} illustrates the prompt for generating the feminine version.


Our prompt is designed such that only the gender-marked words or segments for the speaker are rewritten in feminine and masculine forms, leaving the rest of the translation unaffected. It is also important to note that GPT4 may also be prompted to improve the translations \cite{jiao2023chatgpt} along with gendered translation task but not explored in this work. An ablation study concerning the GPT-4 prompt for gender forms is discussed in Section \ref{sec:llm_results}.


\subsubsection{ Updated ST training targets}
Before reformulation, most translations predominantly use masculine forms, regardless of the speaker’s gender. The reformulation process using GPT-4 yields “gender-debiased” training targets, as depicted in Fig.~\ref{fig:data_prep_fig}. These updated targets are represented in a Gender-Translation format. For instance, (male, masculine) implies that masculine forms are used in translations for male speakers, and so forth. The pairs (male, masculine) and (female, feminine) represent gender-debiased data, where the targets are adjusted according to the input speaker’s gender. Fine-tuning the ST model with this gender-debiased data enables the model to generate gender-specific translations from speech signals. As shown in Fig.~\ref{fig:data_prep_fig}, we also generate (male, feminine) and (female, masculine) pairs using GPT-4, which may initially seem counter-intuitive. However, these pairs are necessary for situations where the gender is predefined and audio cues are not a generalized solution, such as for transgender individuals, non-binary individuals, and children. In these cases, the model should produce the translation based on these predefined gender inputs, without being influenced by audio cues. Furthermore, utterances that do not contain any first-person pronouns in the English transcription are retained as is and referred to as “gender-neutral” data. These data also play a role in our training, as discussed in the following sub-section~\ref{sec:ST_FT}.

\begin{figure}[t]
    \centering
    \vspace*{-0.2cm} 
    \includegraphics[scale=0.6]{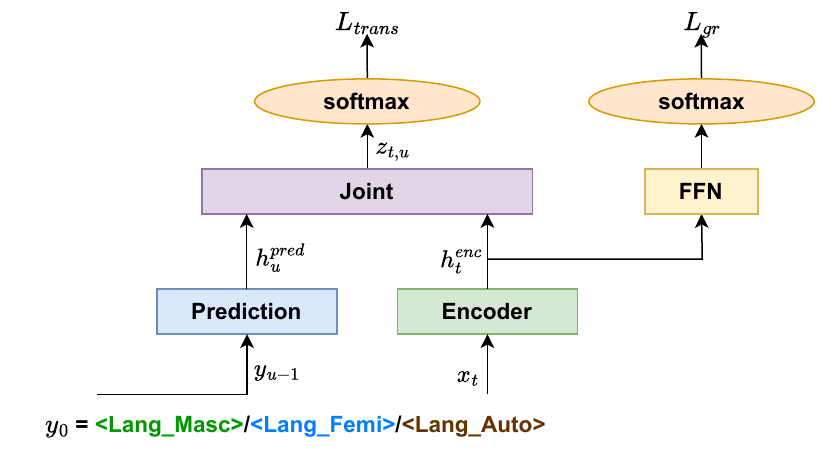}
    \caption{Illustration of model architecture during fine-tuning.}
    \label{fig:model_arch_fig} 
    \vspace*{-0.4cm} 
\end{figure}

\subsection{ST model fine-tuning for gender de-biasing}
\label{sec:ST_FT}
When training ST models with large datasets, retraining the model from scratch using  small amount of gender-debiased data can be computationally expensive and model may also not generalize well. Instead, we propose a more efficient approach: fine-tuning the pre-trained model, denoted as $ST_{Base}$, with gender-debiased data. To mitigate over-fitting, we fine-tune using a combination of “gender-neutral” and “gender-debiased” data. This approach ensures that the BLEU score remains unaffected while enabling the model to learn gender-specific translations. In Section~\ref{sec:grloss_theta}, we delve into the impact of different neutral data-sampling ratios $\theta_{neut}$ during fine-tuning. We also propose a three-mode training when the gender is predefined and the modes are: 
\begin{itemize}

    \item \textit{Masculine mode:} In this mode, the ST model exclusively generates translations for speakers in the masculine form. 
    \item \textit{Feminine mode:}  Here, the ST model produces translations  for the speaker solely in the feminine form.
    \item \textit{Auto mode:}  In Auto mode, the model dynamically adjusts the translation of gender-marked words based on the speaker’s gender, implicitly estimated from the speech signal.

\end{itemize}
The model devoid of any modes and fine-tuned directly with the combination of “gender-neutral” and “gender-debiased” data is termed as the \textit{1-mode FT}, while the model incorporating three modes to also account for user preference is known as the \textit{3-mode FT}. The $ST_{Base}$ model appends $<Lang>$ as start-of-sentence (SOS) token to the output target to denote the target language. To facilitate the training of \textit{3-mode FT} model, we append $<Lang\_Mode>$ as SOS token to the output targets with $Mode$ representing the training mode and $Lang$ denoting the target language as also shown in Fig.~\ref{fig:model_arch_fig}. The training data for each of the modes in \textit{3-mode FT} is presented in Table \ref{fig:data}. Note that all our finetunings uses gender-neutral data to prevent over-fitting or catastrophic forgetting.
\begin{center}
\begin{table}[t]
\caption{ ST fine-tuning data corresponding to 1-mode and 3-mode model in (Gender, Translation) format. }
\centering
\scalebox{0.9}{
\begin{tabular}{|c|c|c|c|}
\hline
\textbf{}&\textbf{Gender-debiased} &\textbf{Gender-neutral}\\
\textbf{}&\textbf{(1 - $\theta_{neut}$)} &\textbf{($\theta_{neut}$)}\\
\hline
\textbf{1-mode FT }&\textbf{(male, masculine),} &\textbf{(male, neutral), } \\
\textbf{}&\textbf{ (female, feminine)} &\textbf{(female, neutral)}\\
\cline{1-3}
\textbf{3-mode FT  }&\textbf{(male, masculine),} & \\
\textbf{ (`Auto' Mode)}&\textbf{(female, feminine)} &\\
\cline{1-2}
\textbf{3-mode FT }&\textbf{(male, masculine),} &\textbf{(male, neutral), }\\
\textbf{ (`Masc' Mode)}&\textbf{(female, masculine )} & \textbf{(female, neutral) }\\
\cline{1-2}
\textbf{3-mode FT}&\textbf{(male, feminine),} & \\
\textbf{(`Femi' Mode)}&\textbf{(female, feminine)} & \\
\hline
\end{tabular}}
\label{fig:data}
\end{table}
\end{center}
\vspace{-1cm}
\subsection{Gender representation loss}
\label{sec:GR}
Our ST model adopts the transformer-transducer architecture \cite{zhang2020transformer, xue2022large} and consists of three components: an encoder network, a prediction network, and a joint network. The encoder transforms input audio features $x_t$ to generate the hidden representations $h_{t}^{enc}$. To encourage our ST model encoder to capture gender information from speech signals, we introduce a gender representation loss, $L_{gr}$, for every utterance, as defined below:
\begin{equation}
L_{gr} = \sum_{i=1}^{T}CE(o_t,g_u)  \nonumber
\end{equation}
where $T$ is the length of input audio feature sequence,  $g_u$ represents the gender label for utterance $u$ and $o_t$ is the gender label predicted at time $t$ for utterance $u$. $o_t$ is obtained as below:
\begin{equation}
o_t = softmax(W^{out}RELU(W^gh_{t}^{enc}+b^g)) \nonumber
\end{equation}
where ${W^{out},W^g, b^g}$ are additional learnable parameters. The combined loss function $L_{comb}$ is a weighted combination of $L_{gr}$ and the transducer loss $L_{trans}$, where $\alpha$ controls the weight of the gender representation loss as shown below:
\begin{equation}
L_{comb} = \alpha L_{gr} + (1-\alpha) L_{trans} \nonumber
\end{equation}
\section{Experimental results}
\subsection{Dataset}
\subsubsection{Training set}
Our research focuses on addressing gender bias in large-scale speech translation (ST) models. Due to the scarcity of such large amount of training datasets in the public domain, we have developed our baseline ST model using an internal corpus. This corpus encompasses roughly 75,000 hours of speech data procured through different sources, which includes approximately 90 million utterances. Within this dataset, where gender labels are available from the source, 68.2\% of the utterances are categorized as male and 31.8\% as female. The original transcripts are in English, and we employ Microsoft’s translation service to render them into two languages: Spanish (ES) and Italian (IT), which serve as our training targets.
\subsubsection{Test set}\label{sec:test_set}
For evaluation, we rely on the publicly available MuST-SHE set~\cite{bentivogli2020gender}  for both ES and IT. For  our hyper-parameter tuning, we restrict ourselves to the utterances labeled as ``dev" in the MuSTC-v1.0-SET column. The evaluation phase is limited to the remaining utterances not labeled as ``dev". Our evaluation metrics include gendered translation accuracy (GTA) and BLEU~\cite{papineni-etal-2002-bleu}. To define GTA, we first establish term coverage (as defined in \cite{gaido2020breeding}), which represents the proportion of gender-marked words annotated in MuST-SHE that are actually generated by the system. GTA measures the proportion of correct gender realizations among the words on which it is measurable. 

 
 \subsection{Experiment details}
 The ST model follows the streaming transformer transducer architecture with 18 transformer blocks, as described in \cite{xue2022large}. 
 The encoder comprises $18$ Transformer blocks, each containing $320$ hidden nodes, $8$ attention heads and $2048$ feed-forward nodes. The prediction network employs $2$ LSTM layers, with each LSTM layer having $1024$ hidden nodes. The joint network is a feed-forward layer with a size of $512$ times the vocabulary. 
 We utilize a combined vocabulary from ES and IT, consisting of $8K$ subword units. The model input consists of $80$-dimensional log-Mel filter-bank features with $25ms$ windows and a $10ms$ shift. The chunk size of the streaming mask for the Transformer blocks is 25. To guide the model in producing translations for a specific language, we prepend a language-specific SOS token $<Lang>$ at the beginning. In our results, we refer to this model as $ST_{Base}$. It serves as the seed model for our fine-tuning experiments aimed at addressing gender bias. 



\subsection{Results}
Table~\ref{tab:GAST_results} presents the results  on the MuST-SHE evaluation set for our Baseline ($ST_{Base}$) and the  proposed approaches: 1-mode and 3-mode fine-tuning. We report the GTA and BLEU score for both Category 1 and Category 2 sets~\cite{bentivogli2020gender} in the MuST-SHE evaluation set. While our primary focus is on improving Category 1, we also provide accuracy metrics for the Category 2 test to avoid regression. 
Our proposed models demonstrate significant improvement over the baseline in GTA for Category-1 feminine  forms across both Spanish and Italian languages. Specifically, the GTA for EN to ES Category-1 feminine forms increased from $10.12\%$ to $87.05\%$, with a minimal regression on masculine form. Similarly, the GTA for EN to  IT Category-1 feminine forms increased from $8.81\%$ to $84.72\%$.While the BLEU score remains similar or exhibits minor regressions, there is a significant improvement in BLEU scores for feminine translations as the translations contained the correct gender form. Interestingly, despite not focusing on Category-2 gender bias, we still see modest improvements due to balanced presence of feminine forms during fine-tuning. The 1-mode fine-tuning outperforms the 3-mode fine-tuning in “Auto” mode, likely due to the latter’s complex training process hindering direct learning from speech cues. For the experiments described above, we set $\theta_{neut}=0.2$ and $\alpha=0.1$ and its tuning is discussed in Section~\ref{sec:grloss_theta}. Additionally, it is important to highlight that the term coverage for all our models exceeds 80\%, surpassing what has been reported in previous works.  \\
The 3-mode model is also designed to generate translations in the selected gendered form, regardless of the input speaker. Consequently, we evaluated the performance of the 3-mode model on the MuST-SHE dataset for Category-1 Masculine and Feminine forms across \textit{all} speakers, corresponding to the chosen modes. The results are summarized in Table~\ref{tab:3mode}.
We achieve average GTA of more than 87.5\% across all the speakers. \\
Finally, we also compare our proposed approach with other large-scale ST models such as Nvidia Canary~\cite{blog}, Meta Seamless M4T~\cite{barrault2023seamless} and prior works~\cite{bentivogli2020gender, gaido2020breeding}, as shown in Table~\ref{tab:comp}. It is important to highlight that the approach by~\cite{ gaido2020breeding} employs an explicit tag to produce gender-specific translations in contrast to the other models listed in Table~\ref{tab:comp} . Furthermore, our comparative analysis is confined to their single-model architecture to guarantee a fair assessment. Notably, our 1-mode FT and 3-mode FT (in ``auto” mode) models exhibit significantly improved performance on MuST-SHE Category 1 feminine forms improving accessibility for female speakers. 
\begin{center}
\begin{table}[t]
\caption{ Comparing Baseline and proposed methods on MuST-SHE test set.}
\vspace{0.5cm}
\centering
\scalebox{0.78}{
\begin{tabular}{|c|c|c|c|c|c|c|}
\hline
\textbf{}&\multicolumn{2}{|c|}{\textbf{Cat1- Masc}}&\multicolumn{2}{|c|}{\textbf{Cat1- Femi}} &\multicolumn{2}{|c|}{\textbf{Cat2}} \\
\cline{2-7}
\textbf{Model} & \textbf{\textit{Acc.}}  & \textbf{\textit{BLEU}} & \textbf{\textit{Acc.}}  & \textbf{\textit{BLEU}} & \textbf{\textit{Acc.}}  & \textbf{\textit{BLEU}} \\
\hline
\multicolumn{7}{|c|}{ \hspace{40 mm} English (EN) to Spanish (ES)}\\ 
\hline
\textbf{STBase}  & 98.55 & 33.8 & 10.12 & 32.6 & 80.54 &  32.3 \\
\textbf{1-mode FT} & 96.86 & 33.1 & \textbf{87.05} & \textbf{36.50}  & 82.1  &  32.1 \\
\textbf{3-mode FT(``Auto")}  & 91.1 & 33.5 & \textbf{81.1} & \textbf{ 35.7} & 83.1 &  32.5 \\
\hline
\multicolumn{7}{|c|}{  \hspace{40 mm}  English (EN) to Italian (IT) }\\ 
\hline
\textbf{STBase}  & 96.61 & 30.08  & 8.81 & 24.67 & 77.1  &  27.2 \\
\textbf{1-mode FT}  & 96.1 & 29.97 & \textbf{84.72 }& 2\textbf{7.2} & 80.2  &  26.95 \\
\textbf{ 3-mode FT(``Auto")}  & 90.4 & 29.75 & \textbf{80.6 } & \textbf{27} & 80.6 &  27.1 \\
\hline
\end{tabular}}
\label{tab:GAST_results}
\end{table}
\end{center}
\vspace{-1cm}

\subsubsection{Impact of GR loss and $\theta_{neut}$}
\label{sec:grloss_theta}
\begin{figure}[htbp]
    \centering
    \includegraphics[scale=0.4]{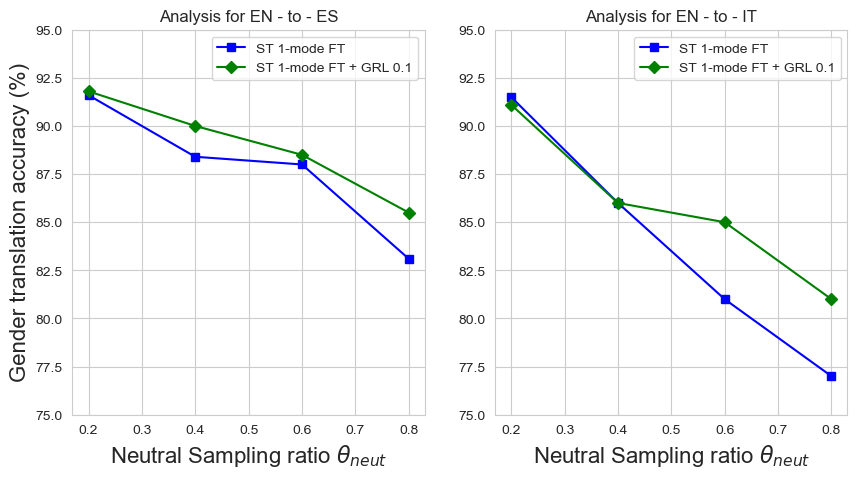}
    \caption{GTA (\%) with and without GR loss for varying $\theta_{neut}$.}
    \label{fig:hyperparameter_tuning}
\end{figure}

Here, we discuss the hyper-parmeter finetuning of neutral data-sampling ratio $\theta_{neut}$. Fig. ~\ref{fig:hyperparameter_tuning} illustrates the gendered translation accuracy (averaged over masculine and feminine forms) with different values of $\theta_{neut}$, both with and without GR loss, on the MuST-SHE \textit{dev} set as described in Section~\ref{sec:test_set}. As $\theta_{neut}$ increases, the accuracy for category-1 feminine forms decreases. However, this reduction in accuracy is accompanied by a reduced risk of overfitting and regression on BLEU scores.
Notably, with $\theta_{neut}<0.2$, we observed a decline in BLEU scores.  Further, the addition of GR loss proves more advantageous at higher values of 
$\theta_{neut}$, thus validating our hypothesis that GR loss enhances the encoder’s ability to capture gender information effectively. Based on above findings, we set $\theta_{neut}=0.2$ and $\alpha=0.1$ for our experiments.

\subsubsection{GPT-4 reformulation analysis}\label{sec:llm_results}
We also assess GPT-4’s ability to reformulate gender forms for reference speakers while not altering the gender form for referent. 
Our evaluation included various prompting strategies: 0-shot, 10-shot (with 10 examples in the prompt), and Chain-of-Thought (providing the reasoning) combined with 10-shot prompting for Spanish (ES) and Italian (IT). The best results are achieved with the 10-shot+CoT approach (Fig.~\ref{fig:GPTprompt}) achieving 94\%  and 93\% accuracy on speaker gender forms for ES and IT respectively  on the MuST-SHE \textit{dev} set as described in Section~\ref{sec:grloss_theta}. We also observed that without CoT, GPT-4 sometimes altered the gender forms for referents, potentially impacting the performance of downstream fine-tuned models on MuST-SHE Category 2 test sets.

\begin{center}
\begin{table}[t]
\caption{ 3-mode FT results for Masculine and Feminine modes}
\vspace{0.3cm}
\centering
\scalebox{0.75}{
\begin{tabular}{|c|c|c|c|c|}
\hline

\textbf{}&\multicolumn{2}{|c|}{\textbf{All Spk / Masc Trans}}&\multicolumn{2}{|c|}{\textbf{All Spk / Femi Trans}}  \\
\cline{2-5}
\textbf{Model} & \textbf{\textit{Acc.}}  & \textbf{\textit{BLEU}} & \textbf{\textit{Acc.}}  & \textbf{\textit{BLEU}} \\
\hline
\multicolumn{5}{|c|}{ \hspace{40 mm} English (EN) to Spanish (ES) }\\ 
\hline

\textbf{3-mode FT(``Masc")}  & 90.2 & 32.9 & - & -    \\
\textbf{ 3-mode FT(``Femi")}  & -  & - & 89.1 & 37.1 \\
\hline
\multicolumn{5}{|c|}{ \hspace{40 mm} English (EN) to Italian (IT) }\\ 
\hline

\textbf{ 3-mode FT(``Masc")}  & 88.2 & 29.6 & - &  -  \\
\textbf{ 3-mode FT(``Femi")}  & - & - & 87.5 &  26.8  \\
\hline
\end{tabular}}
\label{tab:3mode}
\end{table}
\end{center}

\begin{center}
\begin{table}[t]
\caption{ Comparative study of ST models on Category-1 gender bias (\% GTA) on MuST-SHE testset. }
\vspace{0.3cm}
\centering
\scalebox{0.80}{
\begin{tabular}{|c|c|c|c|c|}
\hline
\textbf{Model}&\multicolumn{2}{|c|}{EN - to - ES }&\multicolumn{2}{|c|}{EN - to - IT} \\
\hline
\textbf{}& \textbf{Cat1 }&\textbf{Cat1} & \textbf{Cat1}&\textbf{Cat1}\\
\textbf{}& \textbf{Masc }&\textbf{Femi} & \textbf{Masc}&\textbf{Femi}\\
\hline
\textbf{ Bentivogli et al. \cite{bentivogli2020gender}}  & - & - & 87.19 & 49.53  \\
\textbf{ Gaido et al.\cite{ gaido2020breeding}}  & - & - & 63.72 & 69.83  \\
\textbf{Canary \cite{blog} }  & 98.23 & 10.75  & - & -  \\
\textbf{Seamless M4T \cite{barrault2023seamless}}  &  97.78 & 2.50 & 94.37 & 14.71 \\
\textbf{STBase(Ours)}  & 98.55 & 10.12 & 96.61 & 8.81 \\
\textbf{Proposed- 1-mode FT} & 96.86 & \textbf{87.05} & 96.1 & \textbf{84.72}  \\
\textbf{Proposed- 3-mode FT ( ``Auto") } & 91.1 & \textbf{81.1} & 90.4 & \textbf{80.6 }\\
\hline
\end{tabular}}
\label{tab:comp}
\end{table}
\end{center}
\vspace{-2cm}



\vspace{-1cm}
\section{Conclusions}

Our study presented a novel approach to mitigating speaker gender bias in large-scale speech translation systems, leveraging LLMs for gender-alternative translations. We also proposed three-mode fine-tuning to account for user preference and combined training with gender representation loss. Our methods have demonstrated a significant improvement in translations, particularly for female speakers, marking a substantial advancement over existing systems. Future work may involve exploring bias of other types in large-scale ST systems, reducing bias for non-binary speakers and better fine-tuning approaches.

\bibliographystyle{IEEEbib}
\bibliography{gast}

\begin{thebibliography}{10}

\bibitem{zhang2020transformer}
Qian Zhang, Han Lu, Hasim Sak, Anshuman Tripathi, Erik McDermott, Stephen Koo, and Shankar Kumar,
\newblock ``Transformer transducer: A streamable speech recognition model with transformer encoders and rnn-t loss,''
\newblock in {\em ICASSP 2020-2020 IEEE International Conference on Acoustics, Speech and Signal Processing (ICASSP)}. IEEE, 2020, pp. 7829--7833.

\bibitem{li2019improving}
Jinyu Li, Rui Zhao, Hu~Hu, and Yifan Gong,
\newblock ``Improving rnn transducer modeling for end-to-end speech recognition,''
\newblock in {\em 2019 IEEE Automatic Speech Recognition and Understanding Workshop (ASRU)}. IEEE, 2019, pp. 114--121.

\bibitem{bahdanau2014neural}
Dzmitry Bahdanau, Kyunghyun Cho, and Yoshua Bengio,
\newblock ``Neural machine translation by jointly learning to align and translate,''
\newblock {\em arXiv preprint arXiv:1409.0473}, 2014.

\bibitem{stahlberg2020neural}
Felix Stahlberg,
\newblock ``Neural machine translation: A review,''
\newblock {\em Journal of Artificial Intelligence Research}, vol. 69, pp. 343--418, 2020.

\bibitem{berard2018end}
Alexandre B{\'e}rard, Laurent Besacier, Ali~Can Kocabiyikoglu, and Olivier Pietquin,
\newblock ``End-to-end automatic speech translation of audiobooks,''
\newblock in {\em 2018 IEEE International Conference on Acoustics, Speech and Signal Processing (ICASSP)}. IEEE, 2018, pp. 6224--6228.

\bibitem{gaido2020knowledge}
Marco Gaido, Mattia~A Di~Gangi, Matteo Negri, and Marco Turchi,
\newblock ``On knowledge distillation for direct speech translation,''
\newblock {\em arXiv preprint arXiv:2012.04964}, 2020.

\bibitem{xue2022large}
Jian Xue, Peidong Wang, Jinyu Li, Matt Post, and Yashesh Gaur,
\newblock ``{Large-Scale Streaming End-to-End Speech Translation with Neural Transducers},''
\newblock in {\em Proc. Interspeech 2022}, 2022, pp. 3263--3267.

\bibitem{xu2023recent}
Chen Xu, Rong Ye, Qianqian Dong, Chengqi Zhao, Tom Ko, Mingxuan Wang, Tong Xiao, and Jingbo Zhu,
\newblock ``Recent advances in direct speech-to-text translation,''
\newblock {\em arXiv preprint arXiv:2306.11646}, 2023.

\bibitem{gandhareliterature}
Sanket Gandhare, Preethi Jyothi, and Pushpak Bhattacharyya,
\newblock ``Literature survey: Spoken language translation,''
\newblock {\em Sanket\_SurveyPaper\_SPKMT. pdf}, 2018.

\bibitem{bentivogli2021cascade}
Luisa Bentivogli, Mauro Cettolo, Marco Gaido, Alina Karakanta, Alberto Martinelli, Matteo Negri, and Marco Turchi,
\newblock ``Cascade versus direct speech translation: Do the differences still make a difference?,''
\newblock {\em arXiv preprint arXiv:2106.01045}, 2021.

\bibitem{etchegoyhen2022cascade}
Thierry Etchegoyhen, Haritz Arzelus, Harritxu Gete, Aitor Alvarez, Iv{\'a}n~G Torre, Juan~Manuel Mart{\'\i}n-Do{\~n}as, Ander Gonz{\'a}lez-Docasal, and Edson~Benites Fernandez,
\newblock ``Cascade or direct speech translation? a case study,''
\newblock {\em Applied Sciences}, vol. 12, no. 3, pp. 1097, 2022.

\bibitem{koolen2017these}
Corina Koolen and Andreas van Cranenburgh,
\newblock ``These are not the stereotypes you are looking for: Bias and fairness in authorial gender attribution,''
\newblock in {\em Proceedings of the First Ethics in NLP workshop}. Association for Computational Linguistics (ACL), 2017, pp. 12--22.

\bibitem{sun2019mitigating}
Tony Sun, Andrew Gaut, Shirlyn Tang, Yuxin Huang, Mai ElSherief, Jieyu Zhao, Diba Mirza, Elizabeth Belding, Kai-Wei Chang, and William~Yang Wang,
\newblock ``Mitigating gender bias in natural language processing: Literature review,''
\newblock {\em arXiv preprint arXiv:1906.08976}, 2019.

\bibitem{nadeem2020stereoset}
Moin Nadeem, Anna Bethke, and Siva Reddy,
\newblock ``Stereoset: Measuring stereotypical bias in pretrained language models,''
\newblock {\em arXiv preprint arXiv:2004.09456}, 2020.

\bibitem{savoldi2021gender}
Beatrice Savoldi, Marco Gaido, Luisa Bentivogli, Matteo Negri, and Marco Turchi,
\newblock ``Gender bias in machine translation,''
\newblock {\em Transactions of the Association for Computational Linguistics}, vol. 9, pp. 845--874, 2021.

\bibitem{bentivogli2020gender}
Luisa Bentivogli, Beatrice Savoldi, Matteo Negri, Mattia~Antonino Di~Gangi, Roldano Cattoni, and Marco Turchi,
\newblock ``Gender in danger? evaluating speech translation technology on the must-she corpus,''
\newblock {\em arXiv preprint arXiv:2006.05754}, 2020.

\bibitem{gaido2020breeding}
Marco Gaido, Beatrice Savoldi, Luisa Bentivogli, Matteo Negri, and Marco Turchi,
\newblock ``Breeding gender-aware direct speech translation systems,''
\newblock {\em arXiv preprint arXiv:2012.04955}, 2020.

\bibitem{tatman2017gender}
Rachael Tatman,
\newblock ``Gender and dialect bias in youtube’s automatic captions,''
\newblock in {\em Proceedings of the first ACL workshop on ethics in natural language processing}, 2017, pp. 53--59.

\bibitem{bourguignon2015does}
David Bourguignon, Vincent~Y Yzerbyt, Catia~P Teixeira, and Ginette Herman,
\newblock ``When does it hurt? intergroup permeability moderates the link between discrimination and self-esteem,''
\newblock {\em European Journal of Social Psychology}, vol. 45, no. 1, pp. 3--9, 2015.

\bibitem{Aristar1992GrevilleCG}
Anthony Aristar,
\newblock ``Greville corbett, gender . (cambridge textbooks in linguistics.) cambridge: Cambridge university press, 1991. pp. xix + 363.,''
\newblock {\em Journal of Linguistics}, vol. 28, pp. 542 -- 547, 1992.

\bibitem{saunders-byrne-2020-reducing}
Danielle Saunders and Bill Byrne,
\newblock ``Reducing gender bias in neural machine translation as a domain adaptation problem,''
\newblock in {\em Proceedings of the 58th Annual Meeting of the Association for Computational Linguistics}, Dan Jurafsky, Joyce Chai, Natalie Schluter, and Joel Tetreault, Eds., Online, July 2020, pp. 7724--7736, Association for Computational Linguistics.

\bibitem{saunders-etal-2020-neural}
Danielle Saunders, Rosie Sallis, and Bill Byrne,
\newblock ``Neural machine translation doesn{'}t translate gender coreference right unless you make it,''
\newblock in {\em Proceedings of the Second Workshop on Gender Bias in Natural Language Processing}, Marta~R. Costa-juss{\`a}, Christian Hardmeier, Will Radford, and Kellie Webster, Eds., Barcelona, Spain (Online), Dec. 2020, pp. 35--43, Association for Computational Linguistics.

\bibitem{savoldi-etal-2021-gender}
Beatrice Savoldi, Marco Gaido, Luisa Bentivogli, Matteo Negri, and Marco Turchi,
\newblock ``Gender bias in machine translation,''
\newblock {\em Transactions of the Association for Computational Linguistics}, vol. 9, pp. 845--874, 2021.

\bibitem{hassanawadalla2018gender}
Hany Hassan~Awadalla, Mostafa Elaraby, Ahmed Tawfik, Mahmoud Khaled, and Aly Osama,
\newblock ``Gender aware spoken language translation applied to english-arabic,''
\newblock in {\em 2018 IEEE Proceedings of the Second International Conference on Natural Language and Speech Processing}, February 2018.

\bibitem{savoldi2023test}
Beatrice Savoldi, Marco Gaido, Matteo Negri, and Luisa Bentivogli,
\newblock ``Test suites task: Evaluation of gender fairness in mt with must-she and ines,''
\newblock {\em arXiv preprint arXiv:2310.19345}, 2023.

\bibitem{barrault2023seamless}
Lo{\"\i}c Barrault, Yu-An Chung, Mariano~Coria Meglioli, David Dale, Ning Dong, Mark Duppenthaler, Paul-Ambroise Duquenne, Brian Ellis, Hady Elsahar, Justin Haaheim, et~al.,
\newblock ``Seamless: Multilingual expressive and streaming speech translation,''
\newblock {\em arXiv preprint arXiv:2312.05187}, 2023.

\bibitem{cattoni2021must}
Roldano Cattoni, Mattia~Antonino Di~Gangi, Luisa Bentivogli, Matteo Negri, and Marco Turchi,
\newblock ``Must-c: A multilingual corpus for end-to-end speech translation,''
\newblock {\em Computer Speech \& Language}, vol. 66, pp. 101155, 2021.

\bibitem{kocabiyikoglu2018augmenting}
Ali~Can Kocabiyikoglu, Laurent Besacier, and Olivier Kraif,
\newblock ``Augmenting librispeech with french translations: A multimodal corpus for direct speech translation evaluation,''
\newblock {\em arXiv preprint arXiv:1802.03142}, 2018.

\bibitem{touvron2023llama}
Hugo Touvron, Louis Martin, Kevin Stone, Peter Albert, Amjad Almahairi, Yasmine Babaei, Nikolay Bashlykov, Soumya Batra, Prajjwal Bhargava, Shruti Bhosale, et~al.,
\newblock ``Llama 2: Open foundation and fine-tuned chat models,''
\newblock {\em arXiv preprint arXiv:2307.09288}, 2023.

\bibitem{achiam2023gpt}
Josh Achiam, Steven Adler, Sandhini Agarwal, Lama Ahmad, Ilge Akkaya, Florencia~Leoni Aleman, Diogo Almeida, Janko Altenschmidt, Sam Altman, Shyamal Anadkat, et~al.,
\newblock ``Gpt-4 technical report,''
\newblock {\em arXiv preprint arXiv:2303.08774}, 2023.

\bibitem{ma2023fairness}
Huan Ma, Changqing Zhang, Yatao Bian, Lemao Liu, Zhirui Zhang, Peilin Zhao, Shu Zhang, Huazhu Fu, Qinghua Hu, and Bingzhe Wu,
\newblock ``Fairness-guided few-shot prompting for large language models,''
\newblock {\em arXiv preprint arXiv:2303.13217}, 2023.

\bibitem{wei2022chain}
Jason Wei, Xuezhi Wang, Dale Schuurmans, Maarten Bosma, Fei Xia, Ed~Chi, Quoc~V Le, Denny Zhou, et~al.,
\newblock ``Chain-of-thought prompting elicits reasoning in large language models,''
\newblock {\em Advances in Neural Information Processing Systems}, vol. 35, pp. 24824--24837, 2022.

\bibitem{papineni2002bleu}
Kishore Papineni, Salim Roukos, Todd Ward, and Wei-Jing Zhu,
\newblock ``Bleu: a method for automatic evaluation of machine translation,''
\newblock in {\em Proceedings of the 40th annual meeting of the Association for Computational Linguistics}, 2002, pp. 311--318.

\bibitem{blog}
``Nvidia nemo canary model pushes the frontier of speech recognition and translation,'' \url{https://nvidia.github.io/NeMo/blogs/2024/2024-02-canary/}, 2024.

\bibitem{jia2019leveraging}
Ye~Jia, Melvin Johnson, Wolfgang Macherey, Ron~J Weiss, Yuan Cao, Chung-Cheng Chiu, Naveen Ari, Stella Laurenzo, and Yonghui Wu,
\newblock ``Leveraging weakly supervised data to improve end-to-end speech-to-text translation,''
\newblock in {\em ICASSP 2019-2019 IEEE International Conference on Acoustics, Speech and Signal Processing (ICASSP)}. IEEE, 2019, pp. 7180--7184.

\bibitem{jiao2023chatgpt}
Wenxiang Jiao, Wenxuan Wang, Jen-tse Huang, Xing Wang, Shuming Shi, and Zhaopeng Tu,
\newblock ``Is chatgpt a good translator? yes with gpt-4 as the engine,''
\newblock {\em arXiv preprint arXiv:2301.08745}, 2023.

\bibitem{wang2023document}
Longyue Wang, Chenyang Lyu, Tianbo Ji, Zhirui Zhang, Dian Yu, Shuming Shi, and Zhaopeng Tu,
\newblock ``Document-level machine translation with large language models,''
\newblock {\em arXiv preprint arXiv:2304.02210}, 2023.

\bibitem{papineni-etal-2002-bleu}
Kishore Papineni, Salim Roukos, Todd Ward, and Wei-Jing Zhu,
\newblock ``{B}leu: a method for automatic evaluation of machine translation,''
\newblock in {\em Proceedings of the 40th Annual Meeting of the Association for Computational Linguistics}, Pierre Isabelle, Eugene Charniak, and Dekang Lin, Eds., Philadelphia, Pennsylvania, USA, July 2002, pp. 311--318, Association for Computational Linguistics.

\end{thebibliography}

\end{document}